\documentclass[letterpaper, 10 pt, conference]{ieeeconf}  
\IEEEoverridecommandlockouts                             
\usepackage{amsmath}  
\usepackage{amssymb}
\usepackage{bm} 
\usepackage{graphicx}
\usepackage{subcaption}  
\usepackage{booktabs}
\usepackage{siunitx}
\title{\LARGE \bf
EEG-Driven AR-Robot System for Zero-Touch Grasping Manipulation
}
\author{
Junzhe Wang$^{1}$,
Jiarui Xie$^{2}$,
Pengfei Hao$^{3}$,
Cheng Liu$^{4}$,
Yi Cai$^{5*}$
}
\begin{document}
\maketitle
\thispagestyle{empty}
\pagestyle{empty}
\begin{abstract}
Reliable brain–computer interface (BCI) control of robots provides an intuitive and accessible means of human–robot interaction, particularly valuable for individuals with motor impairments. However, existing BCI–Robot systems face major limitations: electroencephalography (EEG) signals are noisy and unstable, target selection is often predefined and inflexible, and most studies remain restricted to simulation without closed-loop validation. These issues hinder real-world deployment in assistive scenarios. To address them, we propose a closed-loop BCI–AR–Robot system that integrates motor imagery (MI)-based EEG decoding, augmented reality (AR) neurofeedback, and robotic grasping for zero-touch operation. A 14-channel EEG headset enabled individualized MI calibration, a smartphone-based AR interface supported multi-target navigation with direction-congruent feedback to enhance stability, and the robotic arm combined decision outputs with vision-based pose estimation for autonomous grasping.  Experiments are conducted to validate the framework: MI training achieved 93.1\% accuracy with an average information transfer rate (ITR) of 14.8 bit/min; AR neurofeedback significantly improved sustained control (SCI = 0.210) and achieved the highest ITR (21.3 bit/min) compared with static, sham, and no-AR baselines; and closed-loop grasping achieved a 97.2\% success rate with good efficiency and strong user-reported control. These results show that AR feedback substantially stabilizes EEG-based control and that the proposed framework enables robust zero-touch grasping, advancing assistive robotic applications and future modes of human–robot interaction.

\end{abstract}

\section{INTRODUCTION}

Human–robot interaction (HRI) plays a crucial role in assistive robotics, teleoperation, and human-inspired systems  \cite{mohebbi2020human,chellali2010tele,coradeschi2006human}, enabling users to control robots naturally and intuitively. For individuals with severe motor impairments, brain–computer interfaces (BCIs) provide a direct communication pathway between the human brain and external devices without requiring any physical movement  \cite{daly2008brain}. Compared to invasive BCIs, non-invasive BCIs are safer, simpler, and more practical for widespread use  \cite{veena2020review}, although they typically suffer from lower information transfer rates (ITRs)  \cite{larocco2020optimizing}. Among non-invasive BCI modalities, electroencephalography (EEG) is widely adopted for its high temporal resolution, portability, and safety  \cite{soufineyestani2020electroencephalography}. EEG-based paradigms such as steady-state visual evoked potentials (SSVEP) and motor imagery (MI) have demonstrated strong potential for real-time intention decoding in control tasks \cite{piciucco2017steady, ang2016eeg}.

In conventional systems, visual stimuli for EEG-based interaction are often presented on an independent monitor \cite{kosmyna2018attending}. For complex, multi-target tasks, this setup is inconvenient: the display is usually not within the same field of view as the workspace, forcing users to repeatedly shift their gaze to verify task progress. This attention switching degrades the interaction experience and limits the practical usability of BCIs \cite{faller2017feasibility}. To overcome this limitation, augmented reality (AR) has emerged as a powerful tool for enhancing perceptual feedback in robotic control, overlaying virtual elements directly onto the real world \cite{nwagu2023eeg}. Integrating AR into BCIs allows users to select and manipulate virtual representations of physical objects in a more immersive, context-aware manner, reducing cognitive load compared to traditional monitor-based interfaces \cite{gang2018user}.

However, most existing BCI-AR frameworks rely on fixed stimulus–command mappings, lack adaptability to dynamically changing environments, and are often demonstrated only in simplified simulation settings without a direct bridge to physical robotic execution. AR technologies can be classified as video see-through (VST) or optical see-through (OST)  \cite{birlo2022utility}. VST-AR acquires real-world imagery via a camera, fuses it with virtual content, and displays the composite view on a screen, such as a smartphone or tablet  \cite{henderson2011augmented}. OST-AR employs a head-mounted optical combiner to overlay translucent virtual elements directly onto the real-world  \cite{herbeck2024adjusting}, allowing simultaneous perception of both without intermediate video processing \cite{tang2003evaluation}. In this context, VST-AR realized on a handheld device provides practical advantages over OST-based HMDs, primarily for two reasons. First, using an EEG-based BCI headset together with an OST-AR HMD can place excessive pressure on the scalp, causing discomfort during prolonged operation \cite{si2018towards,kosmyna2021assessing}. Second, placing the AR headset directly over EEG electrodes may cause mechanical displacement or uneven contact pressure, degrading signal quality and stability \cite{verwulgen2018determining}. In contrast, a handheld VST-AR device avoids physical interference with EEG sensors, ensuring user comfort and reliable neural signal acquisition.

\begin{figure*}[t]
    \centering
    \includegraphics[width=0.92\linewidth]{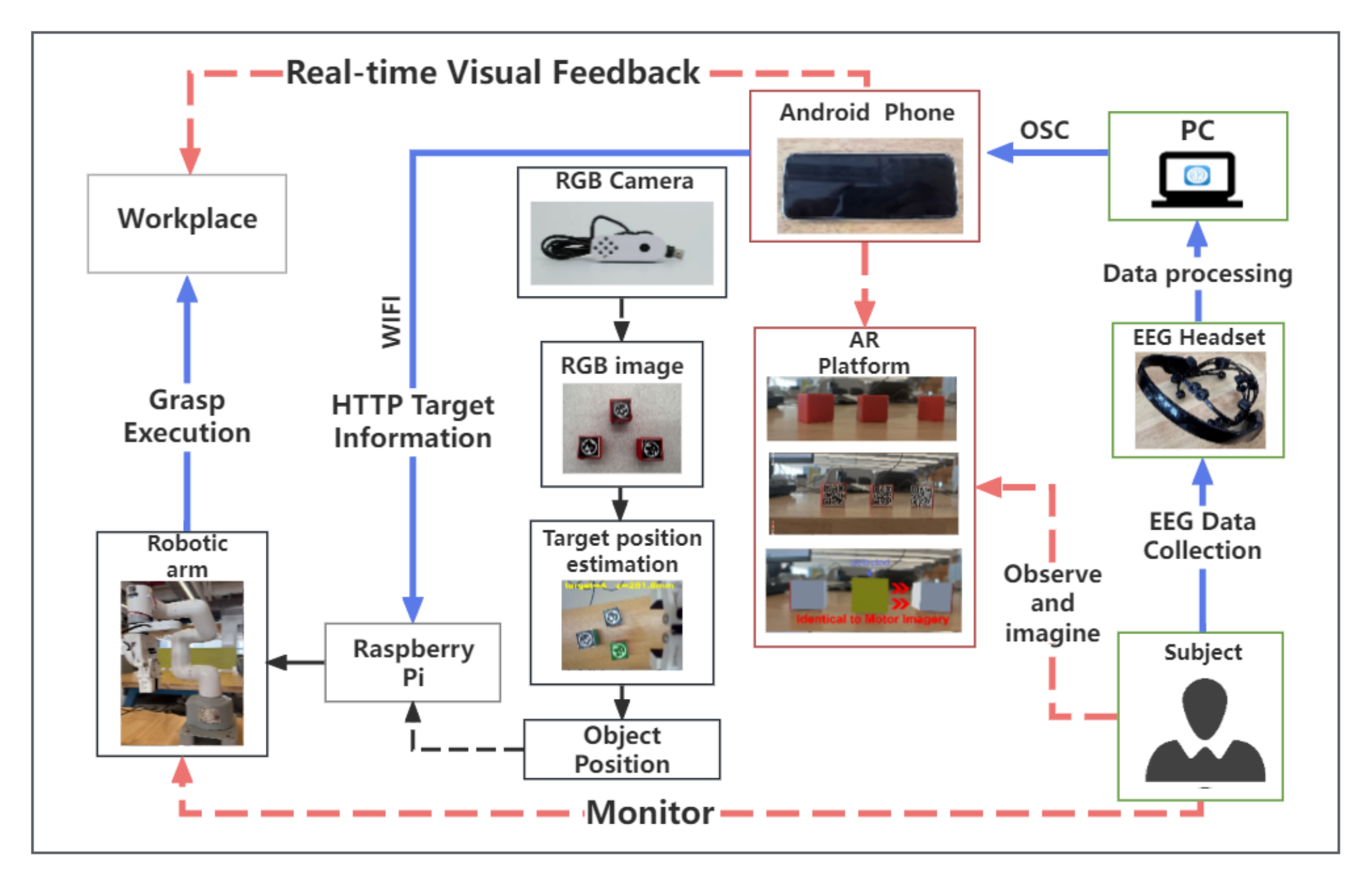}
    \caption{System overview of the proposed BCI–AR–Robot closed-loop grasping framework, integrating (i) EEG-based motor imagery decoding, (ii) AR-based interaction and feedback, and (iii) vision-guided robotic execution.}
    \label{fig:system_architecture}
\end{figure*}

Robotic grasping in complex, multi-target environments poses substantial challenges. EEG signals are inherently noisy, and their effective integration with AR-based real-time visual feedback must be robust to prevent unintended actions (e.g., AR-guided assistance in hybrid gaze–BMI systems has been shown to improve grasping accuracy and reduce user burden significantly \cite{zeng2017closed}). Nevertheless, prior studies on BCI-AR interfaces often rely on predefined and static target layouts, which limit their adaptability to dynamically changing environments \cite{9844133}. Only a few systems demonstrate a complete closed-loop pipeline—from EEG intention decoding, to AR-based target confirmation, to autonomous robotic grasp execution—while leveraging AR feedback to enhance signal persistence and stability. Moreover, recent surveys emphasize the scarcity of fully integrated, physically realistic BCI–AR–robot systems capable of seamless deployment across both physical and simulated settings \cite{ji2023closed, prapas2024connecting, zhang2025mind}.  

Building upon these gaps, we validate a closed-loop framework on the MyCobot 280Pi robot, enabling fully hands-free operation. Users can select and grasp multiple objects solely through EEG control and AR interaction, without the need for manual intervention. The AR feedback mechanism further stabilizes command execution, directly addressing assistive robotics needs by offering an intuitive and accessible interaction modality for mobility-impaired individuals. Experimental results confirm that the system supports flexible multi-object selection, enhances EEG command accuracy, adapts robustly to changing object layouts, and achieves precise grasp execution without requiring manual remapping.

The main contributions of this study are as follows:

\begin{itemize}
    \item \textbf{Integrated framework:} A fully developed multimodal closed-loop pipeline combining EEG intention decoding, AR-based target selection, and autonomous robotic grasping across both physical and simulated environments.
    \item \textbf{Dynamic interaction:} Real-world AR feedback enabling multi-object decisions via EEG commands, ensuring continuity and accuracy in dynamic settings.
    \item \textbf{Robust grasping:} Adaptive strategies based on inverse kinematics control, enhancing stability and transferability across robotic platforms.
    \item \textbf{Hands-free accessibility:} Enabling barrier-free human–robot collaboration, allowing users to accomplish object selection and grasping tasks solely through EEG control, fulfilling requirements for assistive robotics applications.
\end{itemize}

\section{METHOD}

\subsection{System Overview}

We propose a multimodal, closed-loop BCI–AR–Robot system for autonomous grasping. The overall system architecture is illustrated in Fig.~\ref{fig:system_architecture}. The framework addresses three key challenges: robust neural intention decoding, stable target confirmation, and adaptive robotic execution. It integrates four functional modules: 

\begin{itemize}
    \item \textbf{EEG-based intention decoding:} Multiparadigm decoding of MI signals enables direction switching and confirmation commands, allowing users to flexibly navigate and select targets within the AR interface. 

    \item \textbf{AR-based multi-target interaction:} The AR interface dynamically overlays interactive elements onto real-world objects, highlights candidate targets, and maps EEG commands to object selection. A positive feedback mechanism enhances EEG commands' stability and ensures reliable decision-making. 

    \item \textbf{Seamless execution pipeline:} A networked control layer connects the BCI, AR, and robotic arm, supporting real-time communication with improved precision and robustness. 

    \item \textbf{Eye–hand visual perception and grasping:} Robot-mounted cameras detect fiducial markers attached to objects and estimate their poses relative to the robot. Hand–eye calibration aligns the camera and manipulator frames, while inverse kinematics generates feasible grasping trajectories for autonomous execution. 
\end{itemize}

Through this integration, users can complete closed-loop target selection and grasping solely via EEG commands, with AR guidance improving decision efficiency and vision-based calibration ensuring adaptive and transferable robotic operation. The modular architecture also supports scalable deployment across different robotic platforms.

\subsection{BCI and AR Target Selection}

The BCI module employs a motor imagery-based paradigm to generate discrete control commands \{left, right, lift\}, enabling users to navigate across multiple candidate objects and finalize a selection. In the AR interface, each physical object is associated with a fiducial marker, with a virtual interactive block rendered and anchored to the corresponding physical target.

Unlike conventional systems that provide only static responses to EEG commands, our design introduces \emph{Dynamic Visuomotor Neurofeedback}. When the user performs a left or right MI command, the entire interactive AR block, together with its arrow indicator, produces a small but visible sway toward the corresponding neighboring target. This feedback creates the intuitive perception that the imagined movement directly drives the AR environment, thereby reinforcing MI signals, enhancing command persistence, and improving user immersion. When the user issues a lift command, the AR block locks onto the current target object, provides consistent positive feedback, and finalizes the decision-making process.

After confirmation, the AR application transmits the target identifier to the robot side via OSC/HTTP. The robot, equipped with an onboard RGB camera, detects and verifies the corresponding fiducial marker to ensure consistency between AR selection and physical perception.

This interaction design not only enhances user engagement and strengthens the robustness of EEG signals through dynamic, direction-congruent feedback, but also guarantees a fully hands-free workflow in which the entire pipeline—from target selection to grasp execution—is accomplished without manual intervention. This differentiates our framework from prior VR/AR-BCI-Robot systems that lack real-time neurofeedback and direct physical-object coupling. Moreover, it empowers individuals with motor impairments to perform daily tasks more effectively. To rigorously evaluate this innovation, we conducted a comparative user study contrasting static, sham, and dynamic neurofeedback conditions, demonstrating that direction-congruent AR feedback significantly improves command persistence, interaction efficiency, and overall user experience during extended operation.

\subsection{Vision and Eye-in-Hand Calibration}

The experimental platform employs a MyCobot 280Pi, a compact 6-DOF collaborative robotic arm with open SDK support. An eye-in-hand configuration is adopted, with an RGB camera rigidly mounted at the end-effector. The perception module is based on a calibrated pinhole model with intrinsics K and distortion parameters D. Fiducial markers attached to objects are detected in each frame, and their 2D corner observations are matched with the known side length s. The target pose in the camera frame $^{C}\mathbf{T}_{O}$ is then estimated using a Perspective-n-Point solver.

To map detections into the robot base frame, we employ an eye-in-hand calibration model. Offline calibration yields the rigid-body transformation matrix $^{E}\mathbf{T}_{C}$ (camera-to-end-effector). At runtime, the robot provides forward kinematics $^{B}\mathbf{T}_{E}$ (end-effector-to-base). The object pose in the base frame is obtained as:

\begin{equation}
\label{eq:Tchain}
{}^{B}\mathbf{T}_{O} = {}^{B}\mathbf{T}_{E}\, {}^{E}\mathbf{T}_{C}\, {}^{C}\mathbf{T}_{O},
\end{equation}

where $B, E, C, O$  denote the base, end-effector, camera, and object frames, respectively. Calibration quality is quantified by mean reprojection error (in pixels) and repeatability error (in millimeters). To ensure robustness under real-time conditions, pose estimates are stabilized using exponential moving average and median filtering. The fiducial-based approach offers low cost, interference robustness, and industrial reproducibility, making it suitable for real-world deployment.

\subsection{Grasp Synthesis and Execution}

The grasp controller follows a structured waypoint-based execution strategy. Once the target pose $^{B}\mathbf{T}_{O}$ is estimated, the translational component defines the grasp position, while the end-effector retains its current orientation. Four waypoints are synthesized:

\begin{itemize}
  \item {Above pose} $\bigl(T_{\text{above}}\bigr)$: a clearance height above the object $\bigl(+\,Z_{\mathrm{offset}}\bigr)$.
  \item {Approach pose} $\bigl(T_{\text{app}}\bigr)$: a buffered pre-grasp waypoint $\bigl(+\,Z_{\mathrm{approach}}\bigr)$.
  \item {Grasp pose} $\bigl(T_{\text{grasp}}\bigr)$: the compensated contact position, incorporating the gripper-length offset $\bigl(Z_{\mathrm{gripper}}\bigr)$.
  \item {Lift pose} $\bigl(T_{\text{lift}}\bigr)$: a post-grasp validation height $\bigl(+\,Z_{\mathrm{lift}}\bigr)$.
\end{itemize}

Waypoint execution is velocity-adaptive: the robot moves rapidly to
$T_{\text{above}}$, descends at moderate speed to $T_{\text{app}}$, and
approaches $T_{\text{grasp}}$ at low velocity for precision and safety.
After gripper closure, the robot lifts the object to $T_{\text{lift}}$ for
validation. An optional lift offset $Z_{\mathrm{return}}$ ensures obstacle
avoidance before returning to the observation pose.

Error-handling mechanisms are integrated: if reprojection error exceeds a threshold or the fiducial is occluded, the system retries detection or enters a search mode. Unexpected failures during execution trigger a safe-return procedure.

Critically, once the user issues the confirm MI command, the entire pipeline — AR feedback, vision-based localization, and grasp execution — runs fully autonomously without manual adjustment. This hands-free design distinguishes our framework from semi-manual VR/AR-BCI-Robot systems, enabling a complete, real-time closed-loop from neural intention to physical grasp.

\section{EXPERIMENTAL SETUP}

\subsection{Apparatus}
The BCI-AR-Robot control system consists of an EEG acquisition device (Emotiv EPOC X, 14 channels, 128 Hz), an AR interface (a smartphone running a Unity-based AR application), a robotic arm (Elephant Robotics MyCobot 280Pi equipped with a two-finger gripper and an onboard Raspberry Pi), an RGB camera mounted for object perception, and a personal computer (PC) for EEG signal processing. The EEG headset acquires motor imagery signals, the AR interface displays target objects and receives EEG-driven commands, the robotic arm executes grasping tasks via the Raspberry Pi, which performs object pose estimation and kinematic computation, while the PC integrates EEG decoding and transmits high-level control commands to the robotic platform.

\subsection{Participants}
Three healthy adult volunteers (2 males, 1 female, aged 22--24 years) participated in the experiments. All participants had normal or corrected-to-normal vision and reported no history of neurological or psychiatric disorders. None of them had prior experience with BCI-based robot control, ensuring that the system evaluation reflected naïve user performance. All participants gave informed consent prior to the experiment.

\subsection{System Setup}

The system integrates EEG-based MI decoding, AR feedback, and robotic execution into a unified closed-loop pipeline. Prior to online operation, participants complete a calibration phase in which EEG signals corresponding to rest and three MI commands (left, right, lift) are collected and used to train individualized classifiers in EmotivBCI. During real-time use, EEG signals from the EPOC X headset are continuously decoded into discrete commands, driving a Unity-based AR interface on a smartphone for multi-target navigation and confirmation. In this process, virtual interactive objects in the AR environment provide direction-congruent feedback in response to MI commands, thereby enhancing command stability and persistence. Once a target is confirmed, the AR interface transmits the identifier to the robot, where fiducial-based vision estimates the target pose and maps it to the robot base frame. The robot controller then computes the inverse kinematics solution and autonomously executes the grasp before returning to its observation pose.

Through this calibration and integration of neural decoding, AR interaction, and vision-guided manipulation, the system achieves robust, real-time, and fully automated human–robot interaction without manual intervention.

\subsection{Experiment 1: MI Command Training and Calibration}

MI command training and calibration consisted of two stages: Online acquisition and real-time testing. EEG signals were recorded using an Emotiv EPOC X headset (14 channels, 128 Hz) with EmotivBCI 4.8.0317 software. Each participant trained three MI commands \{Left, Right, Lift\}, interleaved with the Neutral state. 

During MI training, participants were instructed to vividly imagine an object in front of them moving toward the corresponding direction (leftward, rightward, or upward/lifting). The Neutral state required participants to remain completely relaxed without performing any mental imagery, serving as a baseline condition. Each trial followed a fixed timing structure: Baseline (2 s), MI imagery (10 s), and Neutral (5 s). For each command, 12-16 repetitions (3 rounds) were performed, with real-time feedback enabled to facilitate user adaptation.

After training, participants completed a 1-3 min validation session in EmotivBCI without AR, during which the live training feedback mode was enabled to assess decoding accuracy and decision time. This procedure ensured that MI commands could be issued reliably and efficiently before integration into AR-based interaction and robotic tasks. 

For signal processing, EEG recordings were first band-pass filtered between 8-16~Hz to isolate motor-related rhythms. Features were extracted as bandpower estimates in the $\mu$ (8-12~Hz) and $\beta$ (12-16~Hz) ranges. A linear classifier implemented in the EmotivBCI software was trained to discriminate among the three MI classes \{Left, Right, Lift\} against the Neutral baseline. During online validation, classification outputs were continuously updated with real-time feedback to assess decoding accuracy, false activation rate, and information transfer rate (ITR).

\begin{figure}[t]
\centering
\begin{subfigure}[t]{0.49\linewidth}
  \centering
  \includegraphics[width=\linewidth]{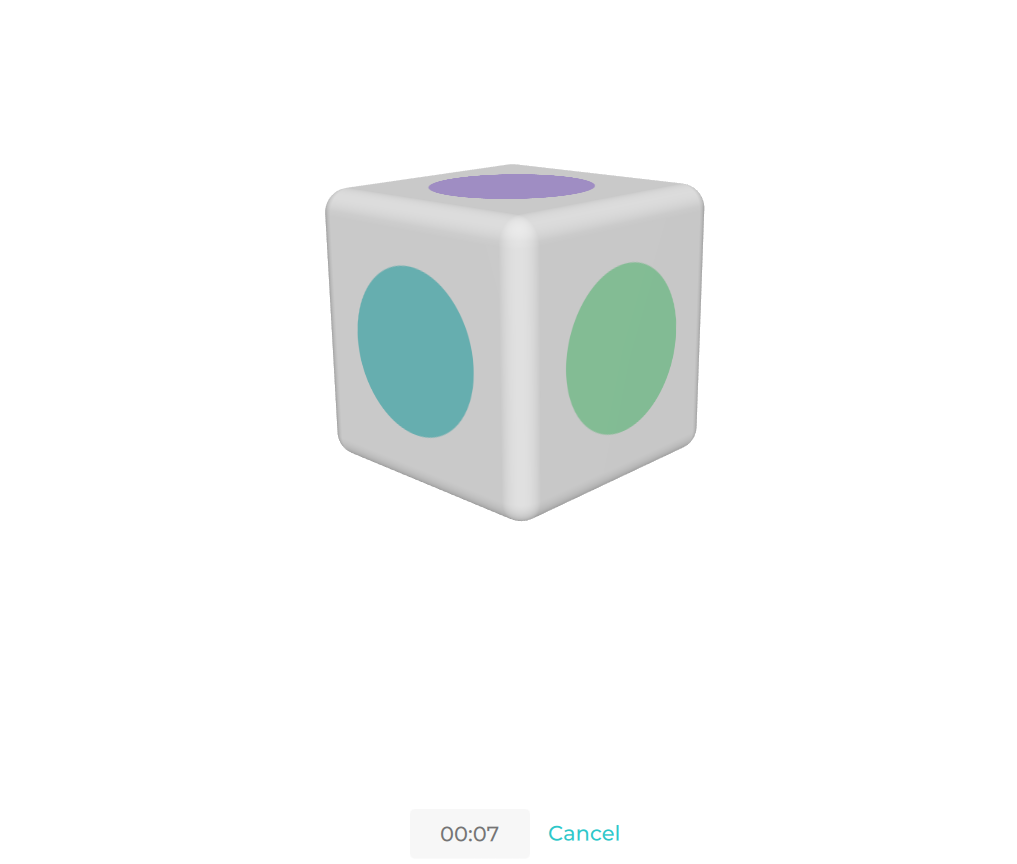}
  \caption{Training preparation phase}
  \label{fig:emotiv_neutral}
\end{subfigure}
\hfill
\begin{subfigure}[t]{0.49\linewidth}
  \centering
  \includegraphics[width=\linewidth]{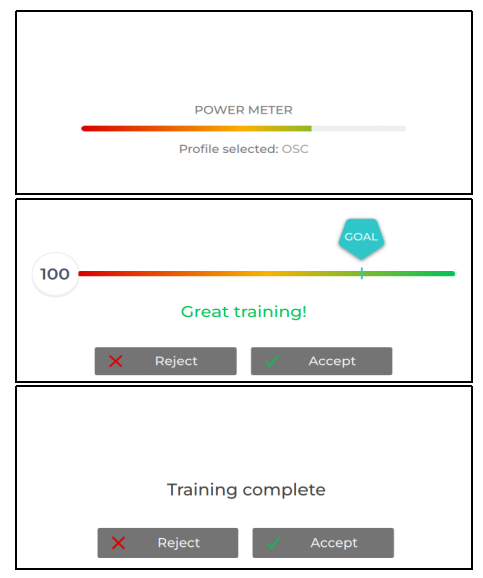}
  \caption{Training progress phases}
  \label{fig:emotiv_left}
\end{subfigure}

\caption{EmotivBCI training and validation interfaces: 
(a) Training preparation; 
(b) During training-End of training (scored)-Saving of training results. }
\label{fig:emotiv_bci}
\end{figure}

\begin{figure}[t]
    \centering
    \includegraphics[width=\linewidth]{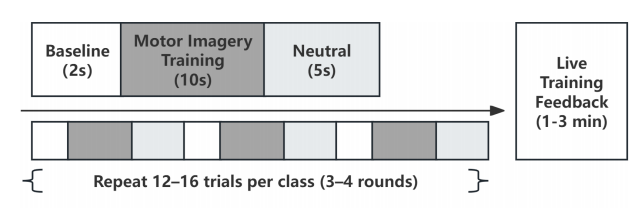}
    \caption{MI training timeline.}
    \label{fig:mi_timeline}
\end{figure}

The complete MI training timeline and interface snapshots are shown in Fig.~\ref{fig:mi_timeline} and Fig.~\ref{fig:emotiv_bci}. Representative raw EEG signals are presented in Fig.~\ref{fig:emotiv_raw}.

\textbf{Evaluation Metrics} 

To quantify the effectiveness of MI training, we report both behavioral and optional neurophysiological measures:

\begin{itemize}
    \item {Accuracy and false activation rate}: the proportion of correctly decoded MI commands and the rate of unintended activations.
    \item {Information Transfer Rate (ITR)}: for $M$ commands, accuracy $P$, and average decision time $T$ (s/selection),
    \[
    \mathrm{ITR} = \Big[\log_2 M + P\log_2 P + (1-P)\log_2\!\Big(\tfrac{1-P}{M-1}\Big)\Big]\cdot \tfrac{60}{T}.
    \]
    \item {Event-Related Desynchronization / Synchronization (ERD/ERS)} (optional): in $\mu$ (8--12 Hz) and $\beta$ (13--30 Hz) bands,
    \[
    \mathrm{ERD\%} = 100 \times \frac{P_{\text{task}} - P_{\text{base}}}{P_{\text{base}}},
    \]
    where negative values (ERD) indicate motor cortex activation and positive values (ERS) reflect post-task rebound.
\end{itemize}

This dual set of measures ensures that both behavioral performance and neural correlates are captured, confirming the robustness of MI commands before integration with the AR-robot system.

\subsection{Experiment 2: Dynamic AR Neurofeedback Ablation}

\begin{figure}[t]
    \centering
    \includegraphics[width=\linewidth]{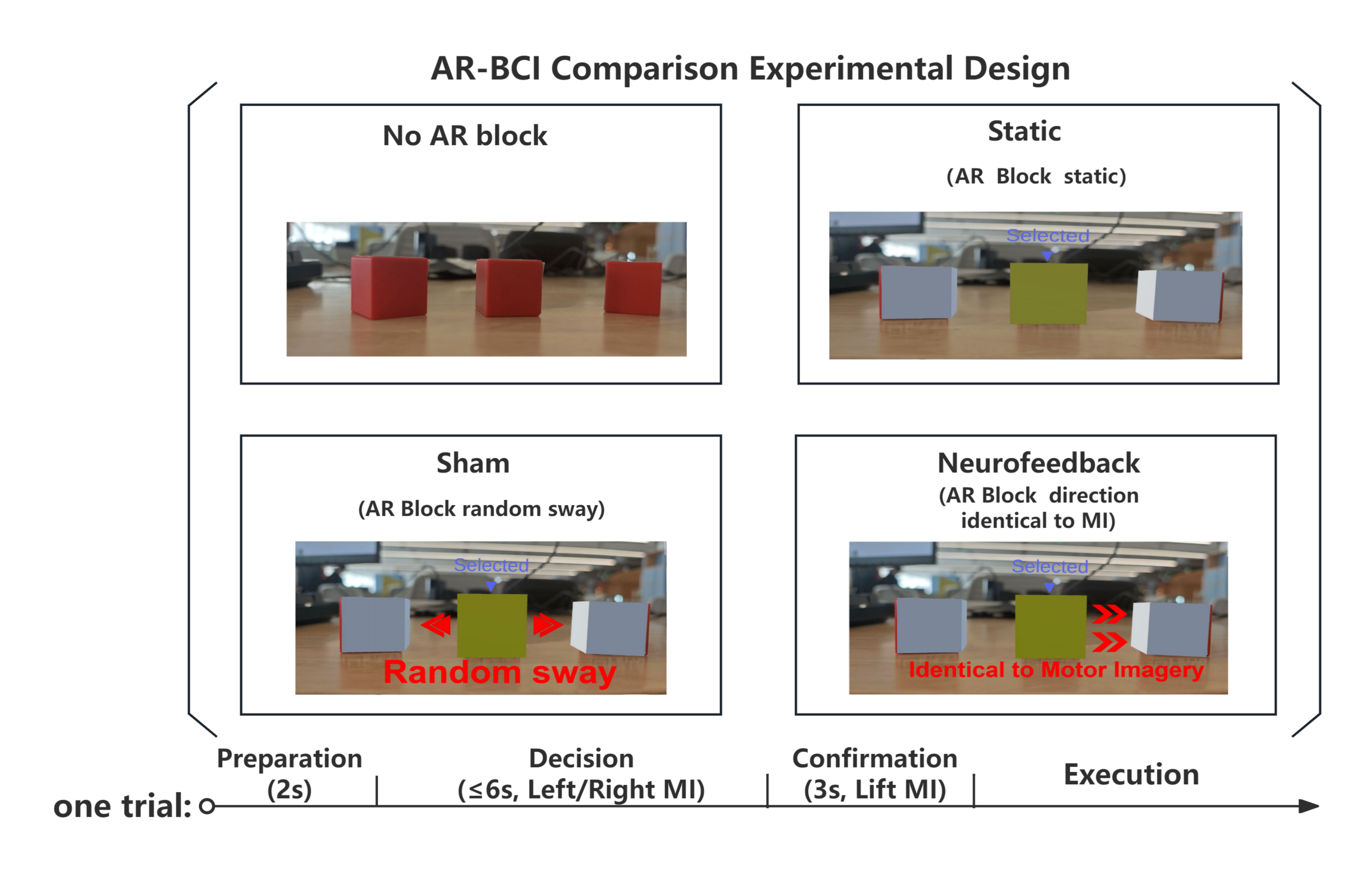}
    \caption{Illustration of Experiment~2. Four BCI-AR feedback conditions and the trial timeline.}
    \label{fig:exp2_conditions}
\end{figure}

We hypothesize that direction-congruent AR neurofeedback enhances the persistence of MI commands and improves interaction efficiency, as reflected in higher ITR and reduced target-switch latency. To test these hypotheses, participants performed trials under four within-subject conditions: (i) \textit{No AR block}, where participants directly issued EEG-based commands to the robot interface without AR visualization; (ii) \textit{Static}, where interactive AR blocks remained fixed and only highlighted upon selection; (iii) \textit{Sham}, where AR blocks swayed laterally with matched amplitude and timing but in directions unrelated to decoded MI outputs; and (iv) \textit{Neurofeedback}, where the AR block exhibits movement in the same direction as the subject's motor imagery direction and switches after a certain period of stabilization (see Fig.~\ref{fig:exp2_conditions}).

Each trial followed a fixed structure: a 2s preparation phase for focusing attention, followed by a decision period lasting up to 6s during which participants switched between candidate targets using left or right MI; this was succeeded by a 3s confirmation phase employing elevated MI to lock onto the selected target; finally, the execution phase commenced, during which robotic movements were disabled to isolate AR feedback effects, with trial success determined by whether target information was successfully transmitted.

\textbf{Evaluation Metrics} 
\begin{itemize}
    \item { \textit{Sustained Control Index (SCI)} }: Quantifies the persistence of MI control over time,
    \[
    \mathrm{SCI} = \tfrac{1}{T} \int_0^T \max\big(0,\, s(t)\cdot d\big)\, dt,
    \]
    where $s(t)\in[-1,1]$ is the signed classifier output and $d\in\{-1,+1\}$ is the cued direction.
    
    \item { \textit{Information Transfer Rate (ITR)} }: Communication efficiency in bits/min, computed as defined in Eq.(1).
    where $M$ is the number of commands, $P$ the classification accuracy, and $T$ the average decision time (s/selection).
    
    \item {  \textit{Average command latency (decision time)} }: Time required to successfully complete a target selection.
    
    \item {  \textit{False Positive Rate (FPR)} }: The proportion of unintended activations relative to the total number of decisions.
\end{itemize}

\subsection{Experiment 3: Closed-Loop Robotic Grasping}

\begin{figure}[t]
    \centering
    \includegraphics[width=0.95\linewidth]{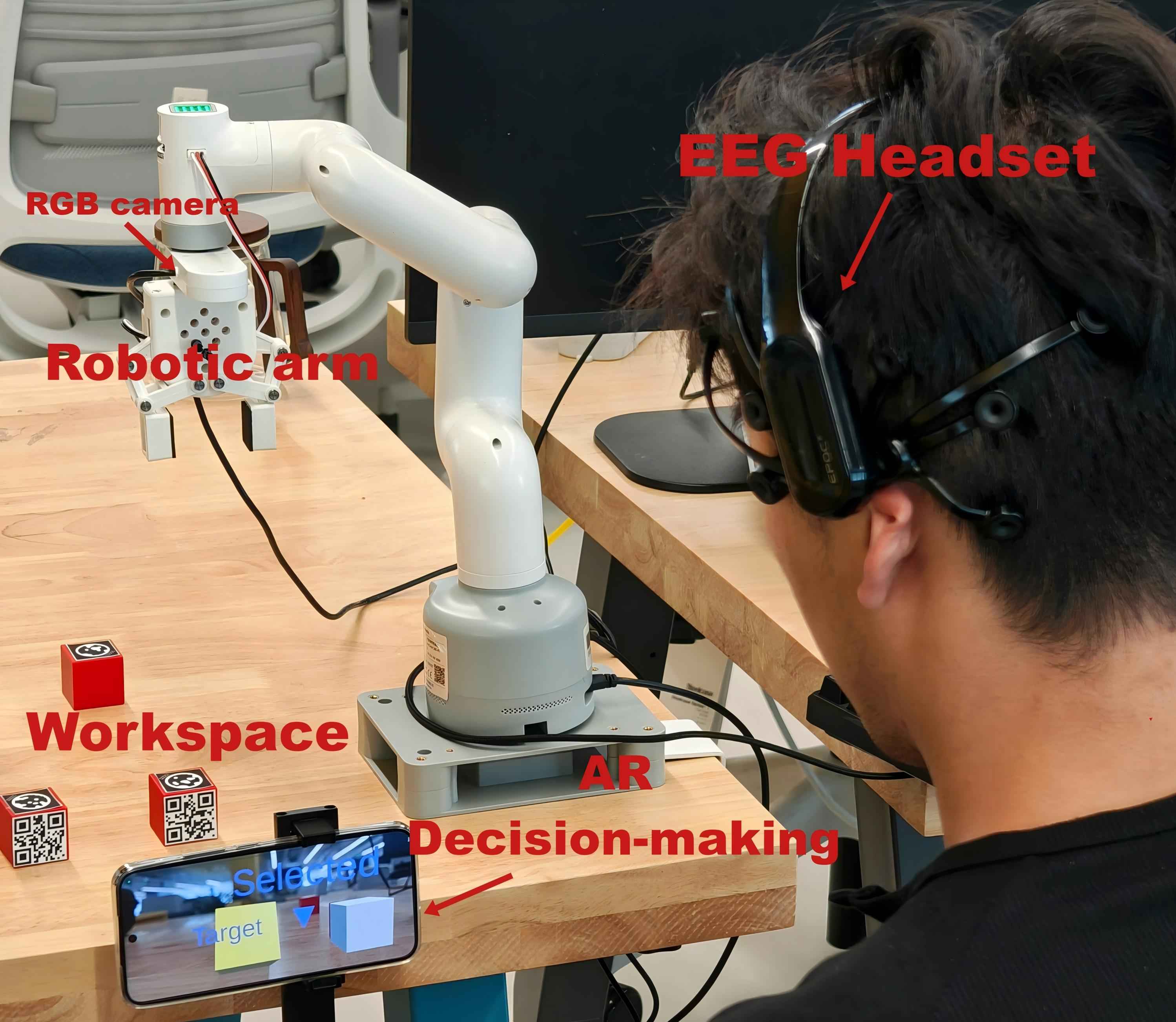}
    \caption{BCI-AR experimental setup for closed-loop robotic grasping.}
    \label{fig:system_setup}
\end{figure}

The third experiment evaluated the complete BCI-AR-Robot pipeline in real-world grasping tasks (Fig.~\ref{fig:system_setup}, Fig.~\ref{fig:system_architecture}). The objective was to validate the accuracy, robustness, and efficiency of the proposed closed-loop system under practical conditions. The experimental setup (Fig.~\ref{fig:system_setup}) shows the subject equipped with an EEG headset, interacting with the AR interface on a smartphone, while the robotic arm with onboard camera executes the grasp. The system architecture (Fig.~\ref{fig:system_architecture}) illustrates the information flow from EEG decoding and AR decision-making to visual perception, inverse kinematics (IK) planning, and robotic execution.

Participants interacted with a $40{\times}40$~cm tabletop workspace containing 3-5 fiducial-tagged objects. Objects were randomly placed but arranged to remain recognizable by the robot’s onboard camera. Using MI commands, participants navigated between AR targets, selected one object as the goal, and confirmed it with the lift command. The robot then autonomously executed the grasp and returned to the observation pose. A trial was considered successful if the object was grasped correctly. Failures occurred if the decision process exceeded the time limit, the grasp failed, or the object could not be recognized.
 
Each participant performed $K=12$ randomized grasping trials. To reduce fatigue, participants rested for 10 minutes after every three trials. For one trial, successful and failed trials were recorded, along with subjective evaluations.

\textbf{Evaluation Metrics} 
\begin{itemize}
    \item {Closed-loop efficiency}: Selection-to-confirm time ($T_{\text{select}}$), confirm-to-grasp planning time ($T_{\text{plan}}$), grasp execution time ($T_{\text{exec}}$), and total cycle time ($T_{\text{total}}$).
    \item {Grasp performance}: Grasp success rate (GSR), placement accuracy (mm), and re-grasp count.
    \item {Robustness}: Performance under layout changes, and lighting variation, with failure cases categorized as EEG misclassification, AR-Robot mapping error, vision failure, or IK failure.
    \item{Subjective assessment}: Workload, sense of agency, and pre/post fatigue ratings.
\end{itemize}

A mixed-effects model was applied with Condition as fixed effect and participant as random effect, followed by Holm-corrected pairwise comparisons against baselines. For censored completion times, Kaplan-Meier survival analysis was used. Ablation studies further removed fiducial markers (vision relying only on texture features) or disabled temporal filtering, quantifying their contributions to grasp success rate and latency.

\section{RESULT}

\subsection{Results of MI Command Training and Calibration}

\begin{figure}[t]
  \centering
  \begin{subfigure}[t]{0.5\linewidth}
    \centering
    \includegraphics[width=\linewidth]{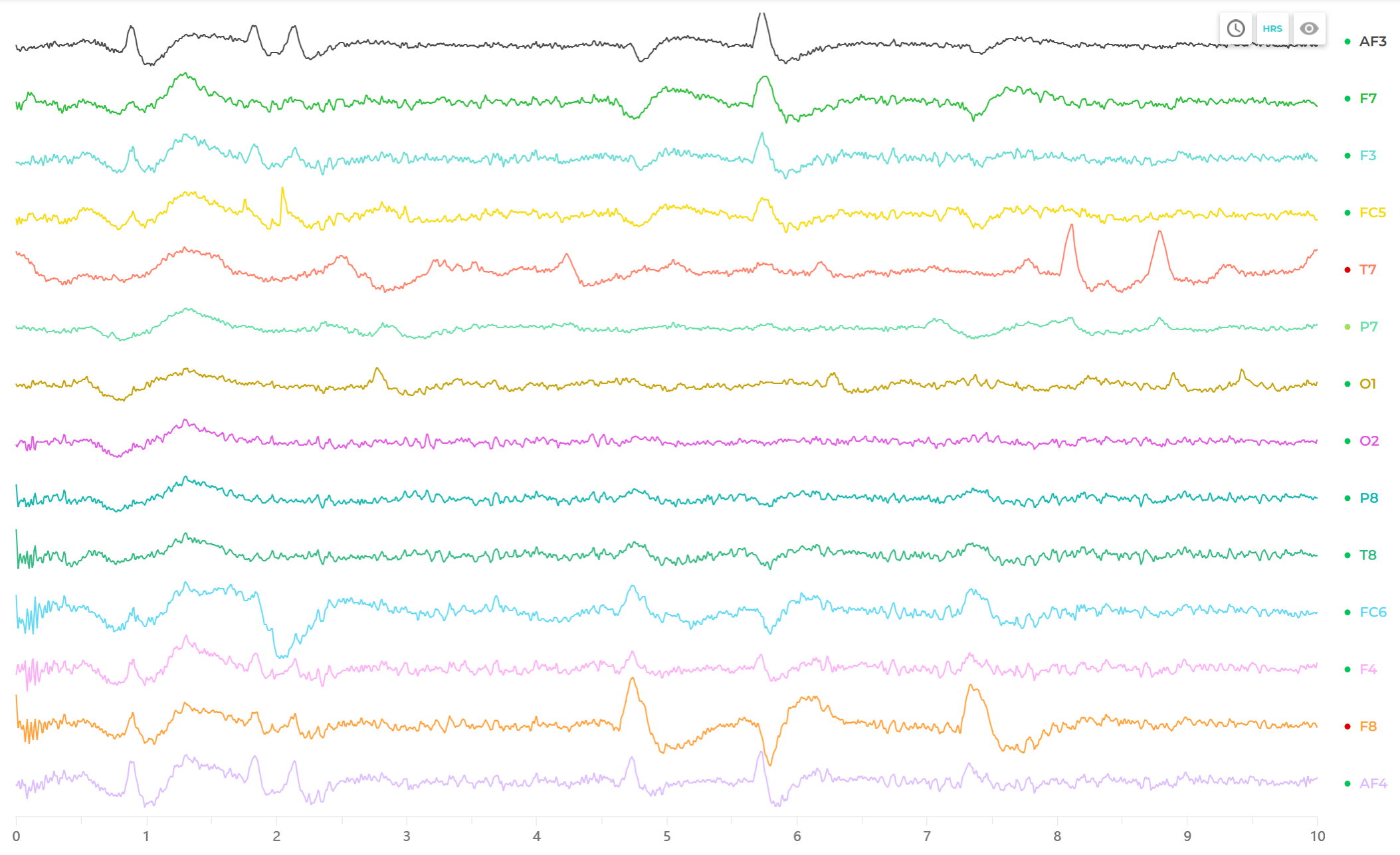}
    \caption{Rest}
    \label{fig:raw_rest}
  \end{subfigure}\hfill
  \begin{subfigure}[t]{0.5\linewidth}
    \centering
    \includegraphics[width=\linewidth]{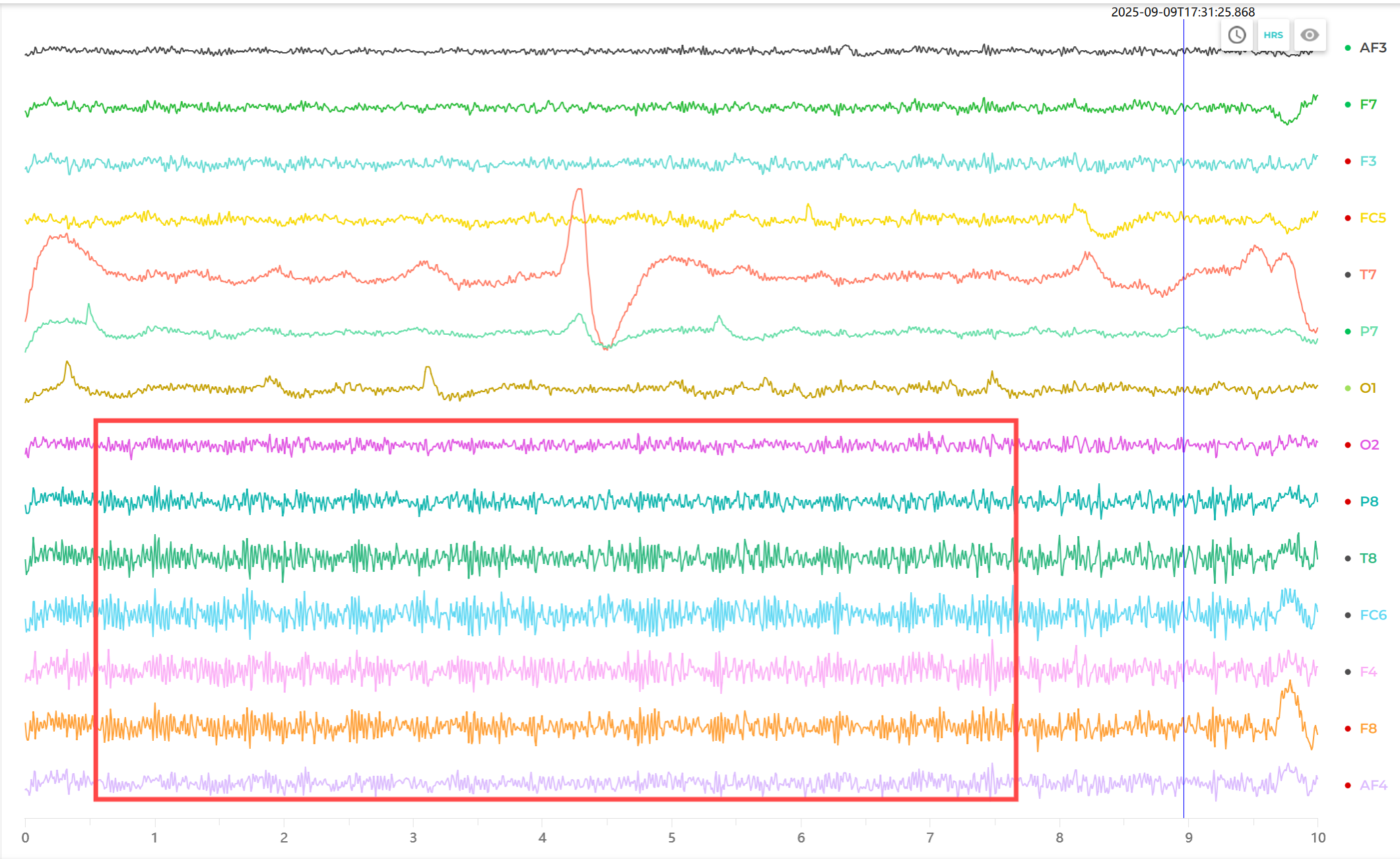}
    \caption{MI: Right.}
    \label{fig:raw_right}
  \end{subfigure}

  \begin{subfigure}[t]{0.5\linewidth}
    \centering
    \includegraphics[width=\linewidth]{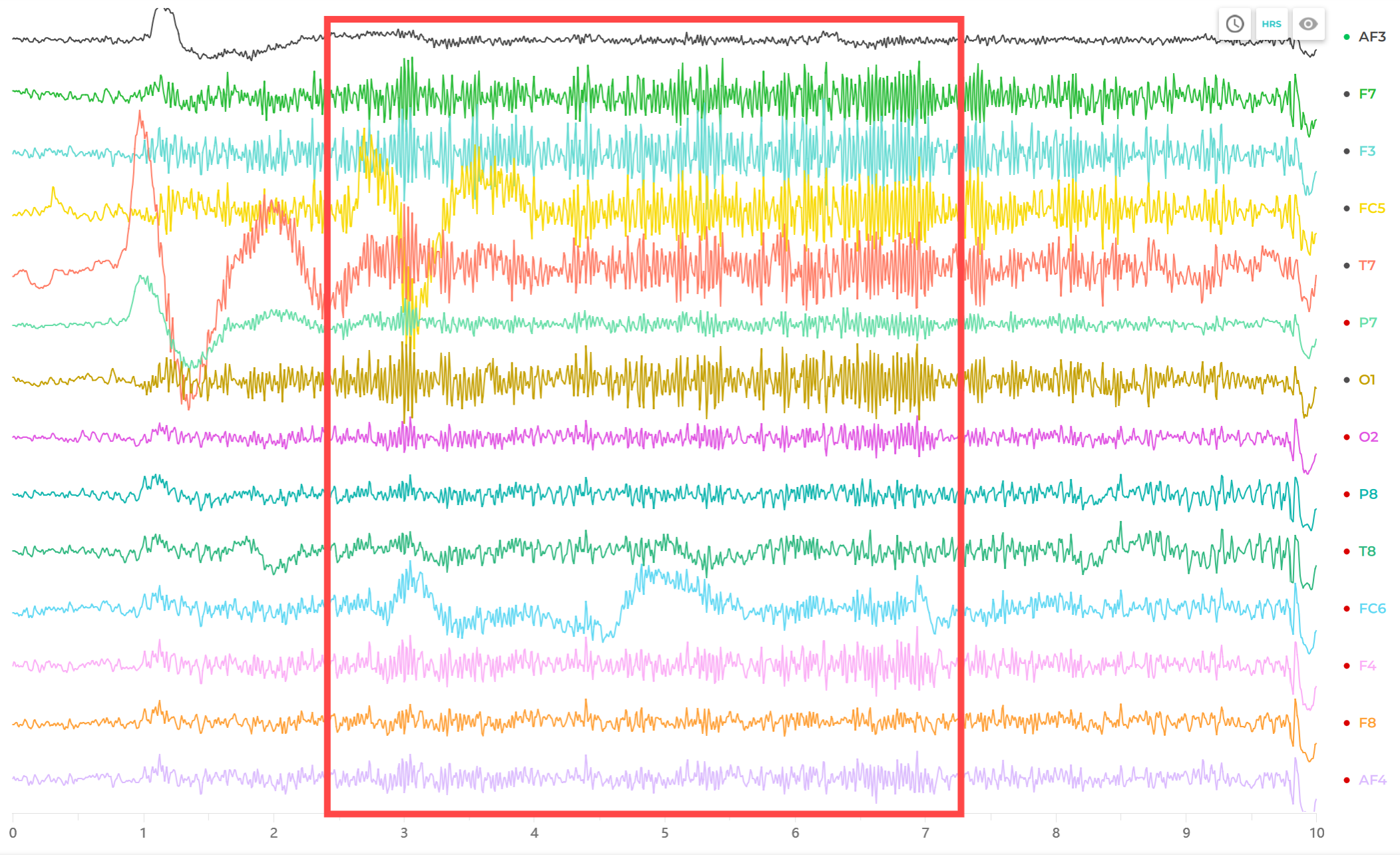}
    \caption{MI: Lift.}
    \label{fig:raw_lift}
  \end{subfigure}\hfill
  \begin{subfigure}[t]{0.5\linewidth}
    \centering
    \includegraphics[width=\linewidth]{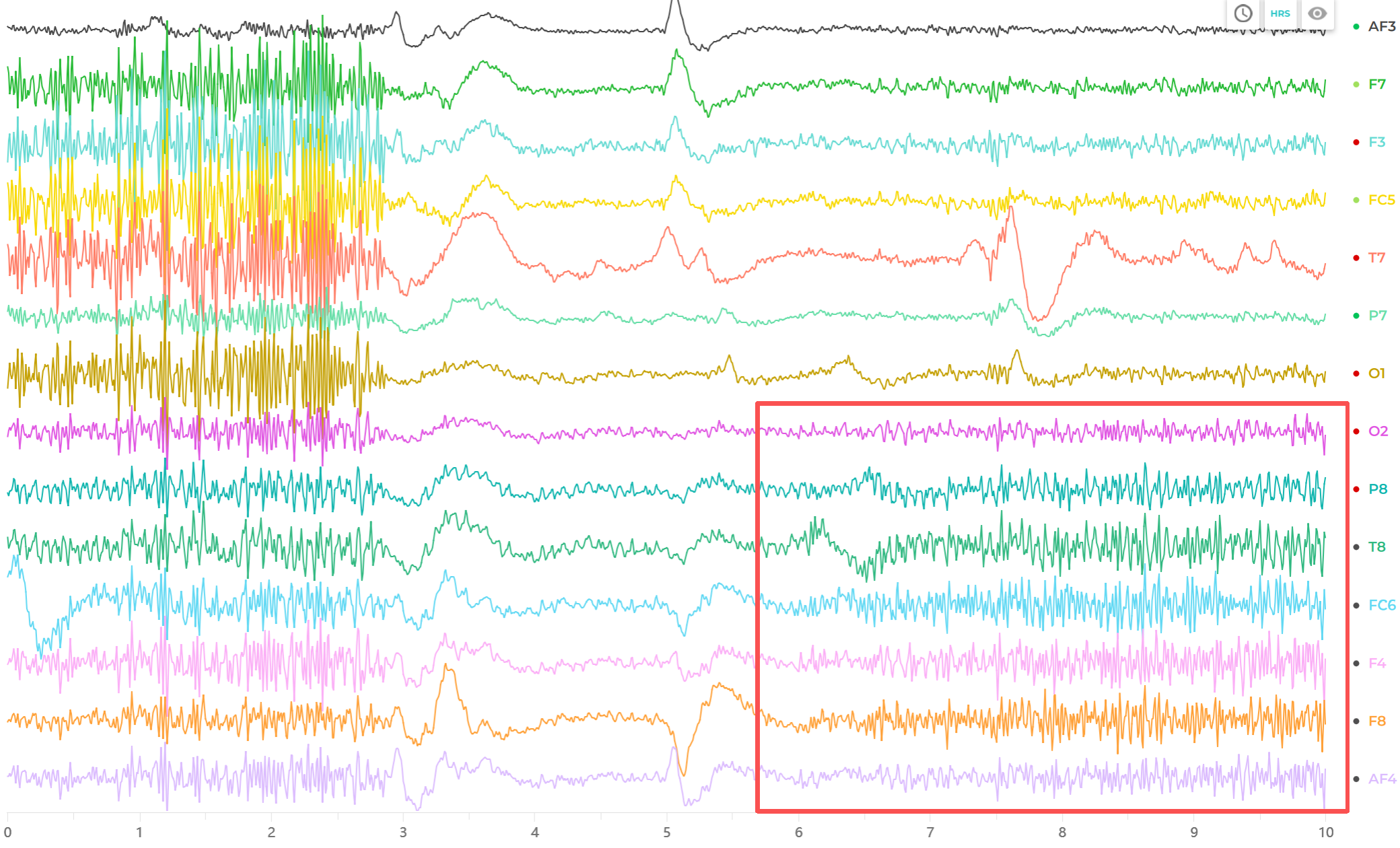}
    \caption{MI: Left.}
    \label{fig:raw_left}
  \end{subfigure}

  \caption{Representative raw EEG across 14 channels under four conditions. Clear task-related fluctuations appear during MI compared with rest.}
  \label{fig:emotiv_raw}
\end{figure}

\begin{figure}[t]
  \centering
  \begin{subfigure}[t]{0.5\linewidth}
    \centering
    \includegraphics[width=\linewidth]{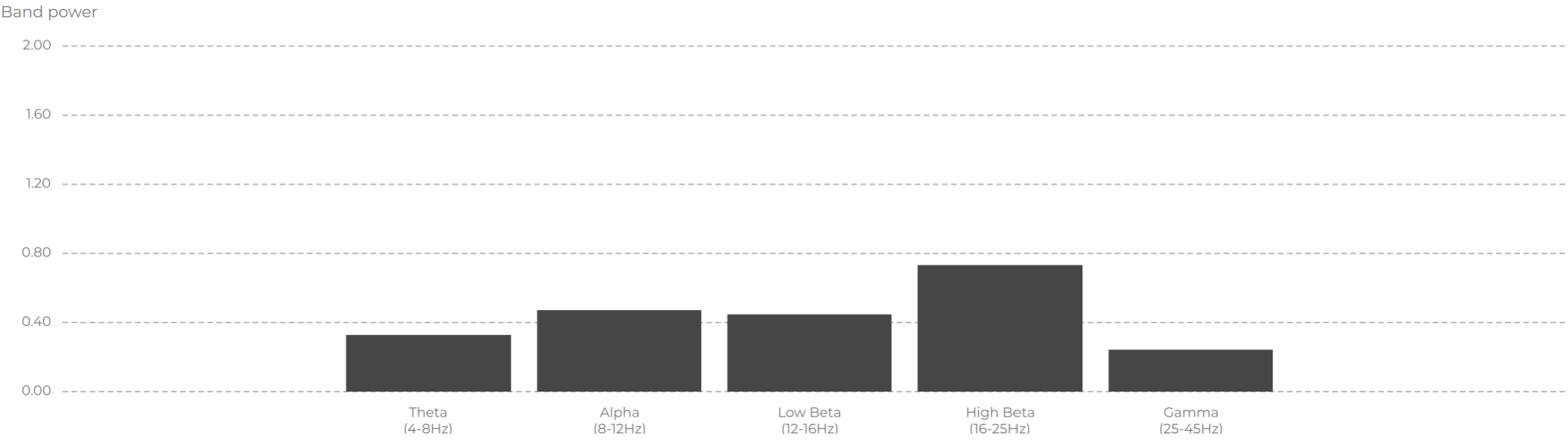}
    \caption{Rest}
    \label{fig:fft_rest}
  \end{subfigure}\hfill
  \begin{subfigure}[t]{0.5\linewidth}
    \centering
    \includegraphics[width=\linewidth]{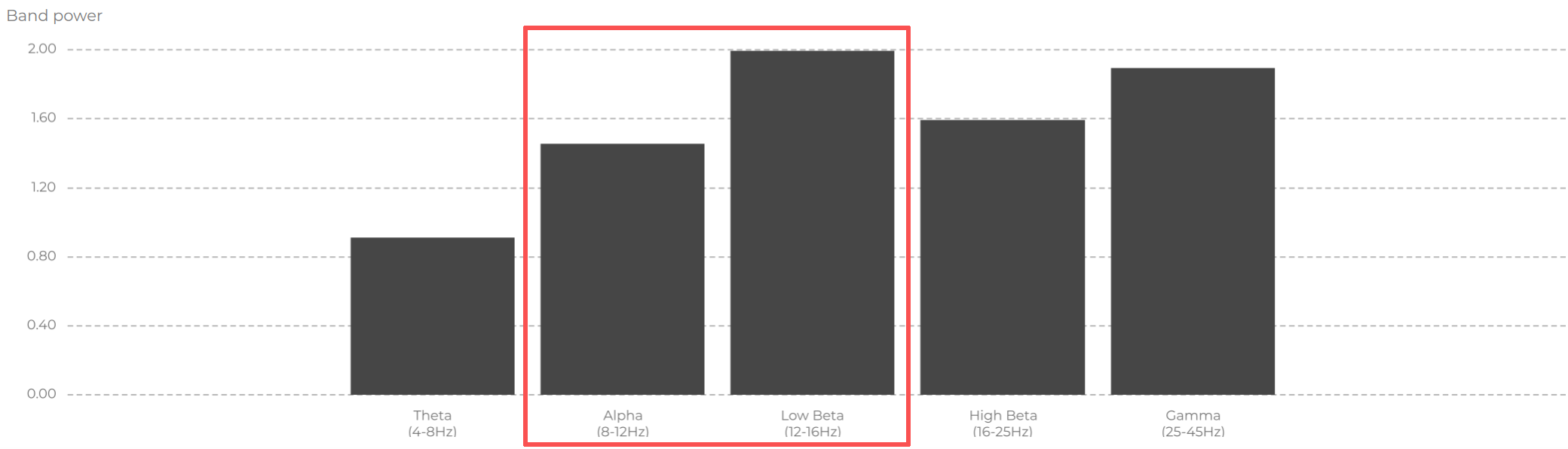}
    \caption{MI: Right.}
    \label{fig:fft_right}
  \end{subfigure}

  \begin{subfigure}[t]{0.5\linewidth}
    \centering
    \includegraphics[width=\linewidth]{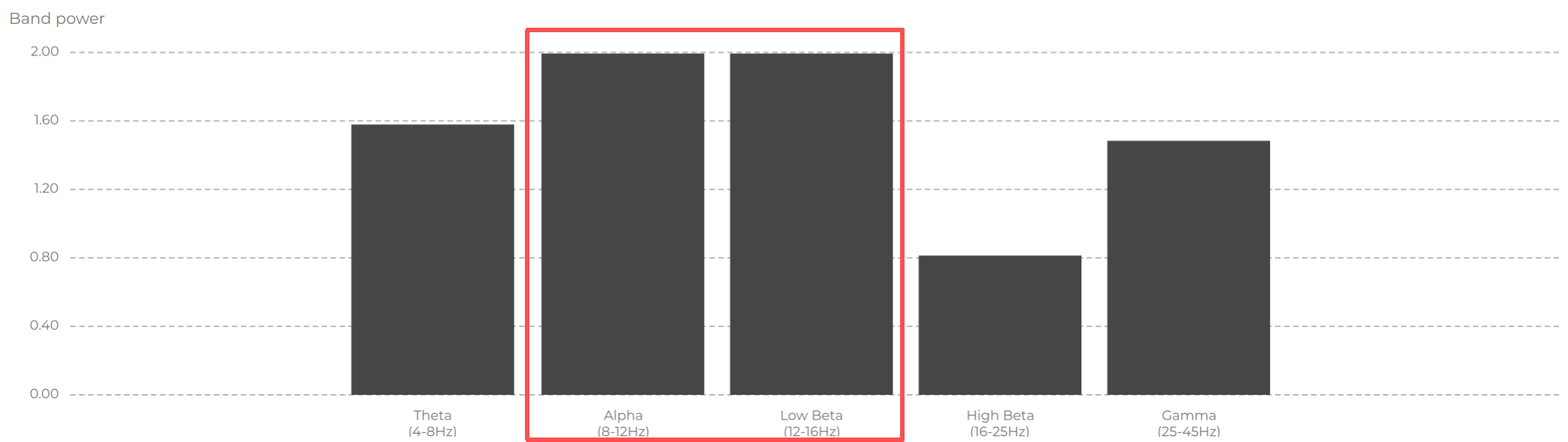}
    \caption{MI: Lift.}
    \label{fig:fft_lift}
  \end{subfigure}\hfill
  \begin{subfigure}[t]{0.5\linewidth}
    \centering
    \includegraphics[width=\linewidth]{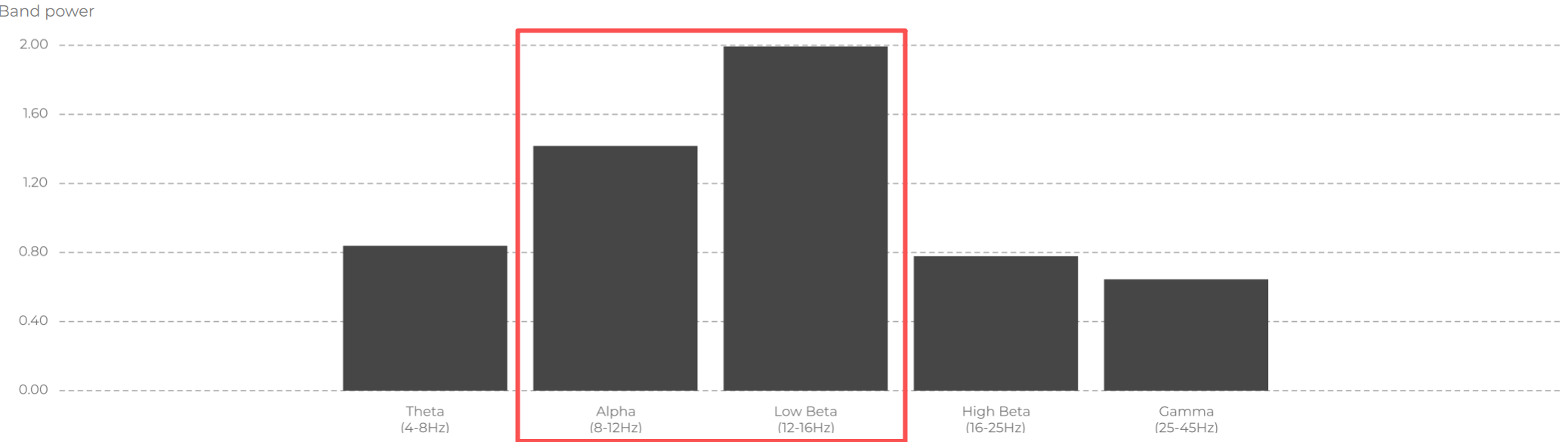}
    \caption{MI: Left.}
    \label{fig:fft_left}
  \end{subfigure}

  \caption{Frequency-domain analysis across four conditions. MI tasks exhibit task-related modulations in the $\mu$ (8--12 Hz) and $\beta$ (12--16 Hz) bands compared with rest.}
  \label{fig:emotiv_spectrum}
\end{figure}

1) EEG Signals Quality and Frequency Features:
Representative raw EEG waveforms across 14 channels during baseline and MI tasks are shown in Fig.~\ref{fig:emotiv_raw}. Compared with the resting state, clear event-related fluctuations emerged during imagery periods, indicating modulation of task-relevant oscillatory activity. In particular, MI tasks elicited distinct changes in the $\mu$ (8--12 Hz) and $\beta$ (12--16 Hz) bands, consistent with motor-related cortical activation. The corresponding frequency-domain analysis (Fig.~\ref{fig:emotiv_spectrum}) further confirmed these task-related modulations, demonstrating that reliable and discriminable neural features were elicited across the three MI conditions.

2) Classification Accuracy and ITR:  
Across participants, the average online decoding accuracy reached 93.1\%, with a false activation rate of 8.3\%. The mean decision time per command was 4.69 $\pm$ 1.20 s. The resulting information transfer rate (ITR) was 14.8 bits/min. Although accuracy varied slightly between individuals, all participants achieved performance above chance level (25\%).

\begin{table}[t]
\centering
\caption{Performance metrics of MI command training and calibration (Experiment~1).}
\label{tab:exp1_results}
\scriptsize
\setlength{\tabcolsep}{3pt}
\begin{tabular}{lcccc}
\toprule
 & Accuracy (\%) & FAR (\%) & Decision Time (s) & ITR (bits/min) \\
\midrule
Group Mean & 93.1 & 8.3 & 4.69 $\pm$ 1.20 & 14.8 \\
\bottomrule
\end{tabular}
\end{table}

\subsection{Results of Dynamic AR Neurofeedback Ablation}

1) Sustained Control and ITR:  
To approximate sustained control, we computed a composite efficiency index (SCI\*) based on accuracy, decision time, and false positive rate. Group-level results revealed that the \textit{Static} (0.223) and \textit{Neurofeedback} (0.210) conditions achieved the highest SCI\*, indicating more stable and efficient motor imagery control compared with \textit{Sham} (0.165) and \textit{No AR} (0.081). As summarized in Table~\ref{tab:exp2_sci}, both \textit{Static} and \textit{Neurofeedback} yielded superior sustained control compared with the non-congruent baselines, while \textit{Neurofeedback} further achieved the greatest ITR (21.3 bits/min), confirming significant improvements relative to all three baselines ($p < 0.05$, Holm-corrected).

\begin{table}[t]
\centering
\caption{Comparison of Sustained Control Index (SCI\*) across 4 AR conditions (Experiment~2).}
\label{tab:exp2_sci}
\scriptsize
\resizebox{0.27\textwidth}{!}{%
\begin{tabular}{lc
}
\toprule
Condition & {SCI\* (mean)} \\
\midrule
No AR         & 0.081 \\
Static        & \textbf{0.223} \\
Sham          & 0.165 \\
Neurofeedback & \textbf{0.210} \\
\bottomrule
\end{tabular}
}
\end{table}

2) Decision Latency and FPR:  
As shown in Table~\ref{tab:exp2_results_full}, the static condition yielded the shortest average decision time (3.76s) but lower accuracy. In contrast, the neural feedback condition achieved a balance between high accuracy and efficient decision-making (4.00s, accuracy = 96.9\%). The no-AR condition resulted in relatively longer and more variable decision times (3.88s) with the lowest accuracy. Although the sham stimulation condition introduced visual motion, the higher false activation rate does not indicate that AR feedback directly affects neural signal quality; rather, it suggests that inconsistent feedback may interfere with users’ control strategies. False positive rates were primarily influenced by EEG signal variability and user adaptation, showing no systematic dependence on AR feedback.
\begin{table}[t]
\centering
\caption{Performance metrics across 4 AR feedback conditions (Experiment 2).}
\label{tab:exp2_results_full}
\scriptsize
\setlength{\tabcolsep}{3pt}
\resizebox{0.5\textwidth}{!}{%
\begin{tabular}{lcccc
               }
\toprule
{Condition} & {Acc. (\%)} & {Time (s)} & {FPR (\%)} & {ITR (bits/min)} \\
\midrule
No AR         & 75.6 & 3.88 & 0.0 & 11.9 \\
Static        & 76.2 & \textbf{3.76} & 0.0 &  9.3 \\
Sham          & 81.0 & 4.16 & 11.1 & 15.7 \\
Neurofeedback & \textbf{96.9} & 4.00 & \textbf{2.8} & \textbf{21.3} \\
\bottomrule
\end{tabular}
}
\end{table}

\subsection{Results of Closed-Loop Robotic Grasping}

1) Closed-Loop Efficiency:  
As summarized in Table~\ref{tab:exp3_all}, the average selection-to-confirm interval ($T_{\text{select}}$) was 15.20 $\pm$ 1.71 s, while planning time ($T_{\text{plan}}$) and robotic execution time ($T_{\text{exec}}$) were 9.44 $\pm$ 0.74 s and 15.26 $\pm$ 0.98 s, respectively. The total cycle time ($T_{\text{total}}$) was 39.90 $\pm$ 1.75 s. Survival analysis of censored trials confirmed that the majority of grasp attempts (97.2\%) were completed within the trial window.

2) Grasp Performance:  
Across three participants, the overall grasp success rate reached 97.2\%, with the only failure attributable to EEG misclassification. No failures were caused by visual occlusion or mechanical errors in the current dataset. These findings confirm that the full BCI-AR-Robot pipeline achieved reliable end-to-end grasp execution.

3) Subjective Assessment:  
Participants reported moderate workload scores on the NASA--TLX, with high ratings for mental demand (81.7~$\pm$~7.6) and effort (81.7~$\pm$~7.6), moderate temporal demand (53.3~$\pm$~5.8), and relatively low physical demand (11.7~$\pm$~7.6). Frustration levels were modest (33.3~$\pm$~15.3).  Sense of agency ratings were high (6.0~$\pm$~1.0 on a 7-point scale), suggesting that participants perceived strong control over the robot. Visual analogue scale (VAS) fatigue ratings increased from pre-task (0.0) to post-task (6.3~$\pm$~0.6), indicating that the experiment induced noticeable but manageable fatigue. 

Overall, these results suggest that the system was cognitively demanding yet provided users with a strong sense of control and an acceptable level of fatigue.

\begin{table}[t]
\centering
\caption{Closed-loop grasping metrics (Experiment~3).}
\label{tab:exp3_all}
\scriptsize
\resizebox{0.43\textwidth}{!}{%
\begin{tabular}{lc}
\toprule
\multicolumn{2}{c}{\textbf{Efficiency (mean $\pm$ SD, s)}}\\
\midrule
$T_{\text{select}}$  & 15.20 $\pm$ 1.71 \\
$T_{\text{plan}}$    &  9.44 $\pm$ 0.74 \\
$T_{\text{exec}}$    & 15.26 $\pm$ 0.98 \\
$T_{\text{total}}$   & 39.90 $\pm$ 1.75 \\
\midrule
\multicolumn{2}{c}{\textbf{Performance outcomes (all trials)}}\\
\midrule
Success rate (GSR)   & 97.2\% \\
Failures (EEG misclassification) & 1 \; (2.8\%) \\
Failures (Vision occlusion)      & 0 \; (0.0\%) \\
Failures (Mechanical error)      & 0 \; (0.0\%) \\
\bottomrule
\end{tabular}
}
\end{table}

\begin{table}[t]
\centering
\caption{Subjective assessment results (Experiment~3). Values are reported as mean $\pm$ SD across participants.}
\label{tab:exp3_subjective}
\scriptsize
\resizebox{0.35\textwidth}{!}{%
\begin{tabular}{lcc}
\toprule
Measure & Mean & SD \\
\midrule
Mental Demand      & 81.7 & 7.6 \\
Physical Demand    & 11.7 & 7.6 \\
Temporal Demand    & 53.3 & 5.8 \\
Effort             & 81.7 & 7.6 \\
Frustration        & 33.3 & 15.3 \\
Sense of Agency (1-7)        & 6.0  & 1.0 \\
Fatigue (VAS)            & 6.3  & 0.6 \\
\bottomrule
\end{tabular}
}
\end{table}

\section{DISCUSSION}

\subsection{Feasibility and Effectiveness}
The experimental results collectively validate the feasibility and effectiveness of integrating MI, AR feedback, and robotic execution into a closed-loop system. Experiment 1 demonstrates that personalized training and calibration enable the reliable decoding of motor imagery commands, with both accuracy and information transfer rates meeting standards, even when using low-channel portable EEG devices. Experiment 2 further reveals that directionally consistent AR neural feedback significantly enhances sustained control and interaction efficiency. Experiment 3 confirmed that the complete BCI-AR-Robot pipeline achieved stable grasping success rates and acceptable closed-loop efficiency in real-world settings. Participants reported moderate workload and a strong sense of autonomous control. These findings validate the system's potential for facilitating efficient and intuitive human-machine interaction. 

\subsection{System Advantages and Innovative Contributions}
The system overcomes limitations of traditional BCI-Robot control by integrating MI decoding with AR feedback, enabling real-time and natural interaction. By combining visual perception with pose estimation, it achieves robust grasping in diverse scenarios. Its closed-loop design unifies neural decoding, AR decision-making, and robotic control, improving command accuracy through AR interaction and minimizing manual intervention. Users can complete target selection and object manipulation entirely hands-free, offering significant benefits for individuals with impaired motor function.  

\subsection{Limitations and Future Work}
This study is limited by the small number of participants, the use of low-channel consumer-grade BCI, and reliance on fiducial markers for vision. Future work will expand participant studies, introduce multimodal signals, develop markerless vision, and investigate adaptive or deep learning-based EEG decoding algorithms to improve robustness. These advances will help extend the system to more complex and dynamic real-world scenarios.

\section{CONCLUSIONS}

 This paper presented a multimodal closed-loop BCI-AR–Robot control framework that integrates motor imagery decoding, augmented reality neurofeedback, and vision-based grasping. Experimental results demonstrated that individualized MI calibration enabled reliable EEG decoding, AR feedback enhanced sustained control and efficiency, and the full system achieved robust grasping performance in real-world tasks. Compared with previous BCI–Robot approaches, the proposed framework introduces ecological AR interaction, real-time seamless BCI–AR–Robot integration, and fully zero-touch operation, offering new perspectives for assistive robotics. Nevertheless, the study is limited by a small participant pool, simplified scenarios, and deterministic grasping strategies; future work will explore broader application methods to enhance the adaptability and generalization of human–robot collaboration.

\addtolength{\textheight}{-12cm}   
\bibliographystyle{IEEEtran}
\bibliography{refs}

\end{document}